\documentclass[lettersize,journal]{IEEEtran}

\usepackage{graphicx}	
\usepackage{amsmath}	
\usepackage{amssymb}	
\usepackage{booktabs}
\usepackage{times}
\usepackage{microtype}
\usepackage{epsfig}
\usepackage[caption=false,font=normalsize,labelfont=sf,textfont=sf]{subfig}
\usepackage{float}
\usepackage{placeins}
\usepackage{color}
\usepackage{colortbl}
\usepackage{stfloats}
\usepackage{enumitem}
\usepackage{makecell}
\usepackage{tabularx}
\usepackage{multirow}
\usepackage{xspace}
\usepackage{url}
\usepackage{xcolor}
\usepackage{arydshln}
\usepackage{cite}
\usepackage{hyperref}

\hypersetup{hypertex=true,
colorlinks=true,
linkcolor=blue,
anchorcolor=blue,
citecolor=blue}

\def\BibTeX{{\rm B\kern-.05em{\sc i\kern-.025em b}\kern-.08em
    T\kern-.1667em\lower.7ex\hbox{E}\kern-.125emX}}

\begin{document}
\title{Large Kernel Modulation Network for Efficient Image Super-Resolution}
\author{Quanwei Hu, Yinggan Tang, Xuguang Zhang
\thanks{Quanwei Hu and Yinggan Tang are with the School of Electrical Engineering, Yanshan University, Qinhuangdao, Hebei 066004, China. Yinggan Tang is also with the Key Laboratory of Intelligent Rehabilitation and Neromodulation of Hebei Province, Yanshan University, Qinhuangdao, Hebei 066004, China. (Email: quanwei1277@163.com; ygtang@ysu.edu.cn)
\par
Xuguang Zhang is with the College of Media Engineering, Communication University of Zhejiang, Hangzhou 310018, China. (Email: 20250002@cuz.edu.cn)
\par
Corresponding author: Yinggan Tang
\par
This work was partly supported by the Hebei Innovation Capability Improvement Plan Project (22567619H).}}

\maketitle
\begin{abstract}
Image super-resolution (SR) in resource-constrained scenarios demands lightweight models balancing performance and latency. Convolutional neural networks (CNNs) offer low latency but lack non-local feature capture, while Transformers excel at non-local modeling yet suffer slow inference. To address this trade-off, we propose the Large Kernel Modulation Network (LKMN), a pure CNN-based model. LKMN has two core components: Enhanced Partial Large Kernel Block (EPLKB) and Cross-Gate Feed-Forward Network (CGFN). The EPLKB utilizes channel shuffle to boost inter-channel interaction, incorporates channel attention to focus on key information, and applies large kernel strip convolutions on partial channels for non-local feature extraction with reduced complexity. The CGFN dynamically adjusts discrepancies between input, local, and non-local features via a learnable scaling factor, then employs a cross-gate strategy to modulate and fuse these features, enhancing their complementarity. Extensive experiments demonstrate that our method outperforms existing state-of-the-art (SOTA) lightweight SR models while balancing quality and efficiency. Specifically, LKMN-L achieves 0.23 dB PSNR improvement over DAT-light on the Manga109 dataset at $\times$4 upscale, with nearly $\times$4.8 times faster. Codes are in the supplementary materials. The code is available at \href{https://github.com/Supereeeee/LKMN}{https://github.com/Supereeeee/LKMN}.
\end{abstract}

\begin{IEEEkeywords}
Efficient image super-resolution, Large convolutional kernel, Feature modulation.
\end{IEEEkeywords}

\section{Introduction}
\IEEEPARstart{I}{mage} super-resolution (SR) refers to the task of reconstructing high-resolution (HR) images from low-resolution (LR) inputs. As a mathematically ill-posed problem, SR has greatly benefited from deep learning techniques such as convolutional neural networks (CNNs) and Transformers, which offer powerful feature extraction and representation capabilities \cite{He2016,Liang2021}. Compared to traditional methods, these approaches have achieved substantial improvements in SR performance. However, such gains often come at the cost of increasingly complex architectures, resulting in a significant rise in model parameters, computational cost, and inference latency. These limitations pose serious challenges for deploying SR models on latency-sensitive edge devices \cite{Ren2024}. To address this issue, developing lightweight and efficient SR models has become a crucial research focus.
\begin{figure}[t]
\centering
\includegraphics[width=1\columnwidth]{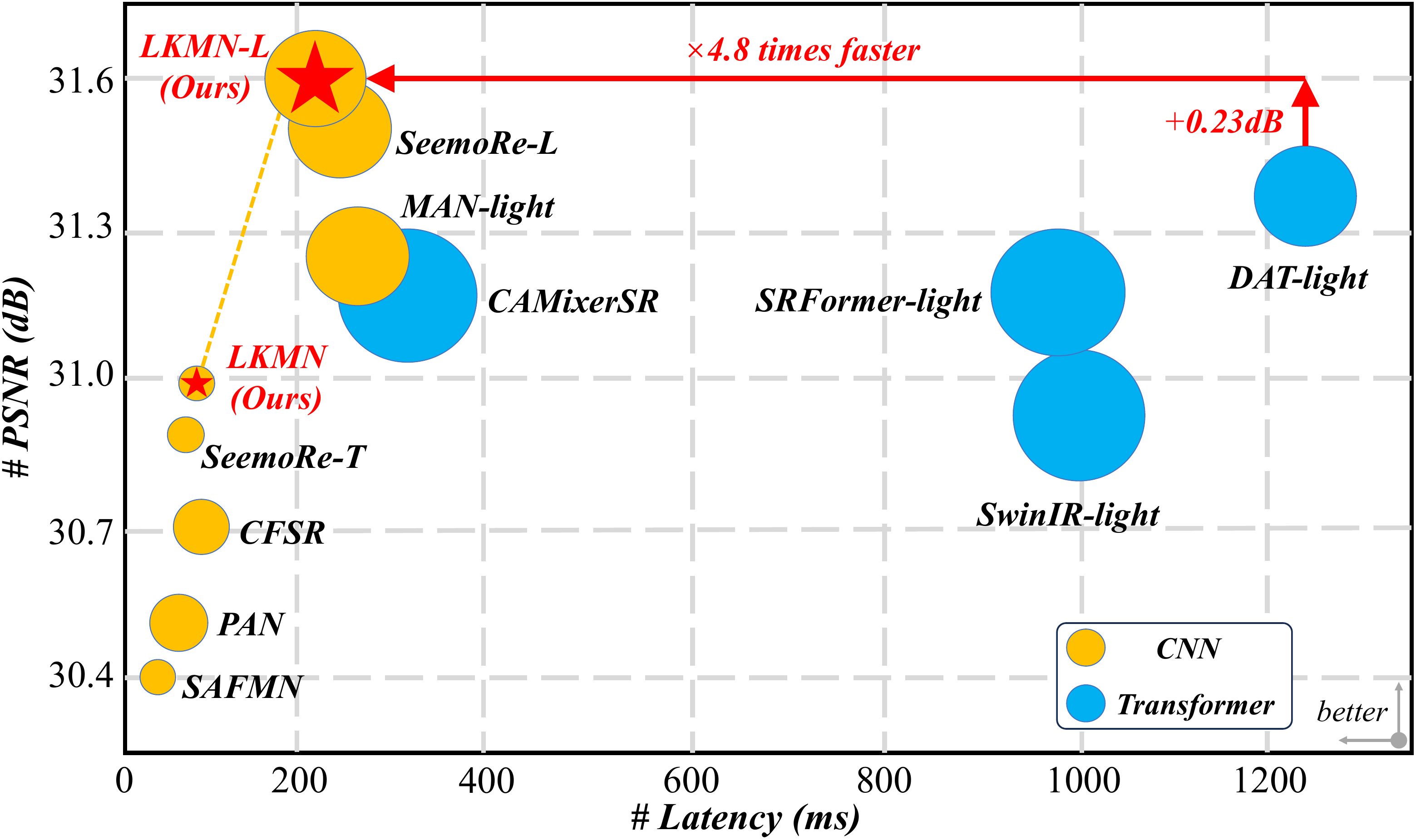}
\caption{Overall SR performance and inference latency comparison between the proposed LKMN and other SOTA lightweight models on Manga109 at $\times$4 upscale. The circle sizes indicate the number of FLOPs.}
\label{fig1}
\end{figure}
\par
Transformer-based efficient SR models, such as SwinIR-light \cite{Liang2021} and SRFormer-light \cite{Zhou2023}, benefit from the powerful long-range modeling capability of the Self-Attention mechanism \cite{Dosovitskiy2021}, thereby achieving SR performance far surpassing that of CNN models. This demonstrates that non-local features are crucial for improving the performance of efficient SR models. However, the Self-Attention mechanism significantly degrades the inference speed of models, which is unacceptable for edge devices.
\par
Currently, CNN-based efficient SR methods like RFDN \cite{Liu2020} and BSRN \cite{Li2022} tend to stack small convolution kernels (3$\times$3) to enhance local feature extraction and minimize latency. Nevertheless, this approach lacks the ability to capture non-local feature information, thus limiting model performance. To address this, methods such as VapSR \cite{Zhou2022} and LKDN \cite{Xie2023} have considered adopting large kernel decomposition and dilated convolutions to expand the receptive field of model, but this inevitably leads to information loss. Furthermore, PLKSR \cite{Lee2024} proposes to directly employ 17$\times$17 large kernel convolutions for non-local feature extraction. While this effectively captures information, it results in a sharp increase in model parameters and computational overhead.
\par
Therefore, how to retain the low-latency advantage of CNNs while acquiring the non-local feature capture capability of Transformers is key to advancing the development of efficient SR. To tackle this challenge, we propose a pure CNN-based model called Large Kernel Modulation Network (LKMN), which comprises two core modules named Enhanced Partial Large Kernel Block (EPLKB) and Cross-Gate Feed-Forward Network (CGFN). The EPLKB employs 31$\times$31 large convolution kernels on partial channels to extract non-local features. Unlike PLKSR, our approach enhances channel information interaction through channel shuffle and channel attention, while further reducing model complexity by decomposing the convolution kernel into horizontal and vertical strip convolutions. The CGFN dynamically adjusts the feature discrepancies among input features, non-local features, and local features, and adopts a cross-gate strategy to enhance the model's capability of modeling and fusing non-local and local features. Based on the aforementioned design, the proposed LKMN achieves superior SR performance compared to other SOTA lightweight SR methods while maintaining comparable model complexity and inference speed.
\par
As shown in Fig. \ref{fig1}, the proposed model strikes a better trade-off between SR performance and inference latency compared with other lightweight SR methods. In summary, our contributions can be summarized as follows:
\begin{itemize}
\item We introduce EPLKB, which employs channel shuffling and partial channel large kernel strip convolution. This design not only retains the non-local feature extraction capability inherent to large-kernel convolution but also effectively reduces model complexity.
\item We propose CGFN, which enables more explicit modeling of the complementarity between local and non-local features through dynamically adjusting their feature discrepancies and incorporating cross-gate strategy.
\item Extensive experiments demonstrate that the proposed LKMN outperforms other lightweight SR methods both quantitatively and qualitatively, while maintaining faster inference speed.
\end{itemize}
\section{Related Work}
\subsection{Efficient Image Super-Resolution}
The introduction of SRCNN \cite{Dong2014} has established CNNs as an effective approach to addressing efficient image SR. For instance, IMDN \cite{Hui2019} and RFDN \cite{Liu2020} employ feature distillation strategies to refine features efficiently while eliminating redundant information. BSRN \cite{Li2022} and SMFANet \cite{Zheng2024} leverage blueprint separable convolution  (BSConv) and depth-wise convolution (DWConv), respectively, to further reduce model complexity. EARFA \cite{Zhao2024} constructs an attention mechanism based on differential entropy to evaluate the importance of channel features. Given that CNNs struggle to capture long-range feature dependencies, Transformer-based SR methods have demonstrated superior performance. SwinIR \cite{Liang2021} reduces computational complexity by computing attention matrices within local windows. SRFormer \cite{Zhou2023} constructs window-based transposed self-attention, thereby achieving more remarkable performance. CAMixerSR \cite{Wang2024a} combines convolution with deformable window self-attention, significantly lowering inference latency. ASID \cite{Park2025} introduces an attention-sharing feature distillation framework, which effectively mitigates the efficiency bottleneck of self-attention. Nevertheless, existing Transformer-based SR methods still lag considerably behind CNN-based ones in terms of inference speed, making them unsuitable for low-computing-power devices with real-time requirements.
\subsection{Large Kernel Network}
Recent research has underscored the effectiveness of large convolution kernels. For instance, SegNeXt \cite{Guo2022} enhances its ability to capture multi-scale contextual information by fusing deep strip convolution branches with large kernel sizes of 7, 11 and 21. PLKSR \cite{Lee2024} employs 17$\times$17 large convolution kernels on partial channels to approximate the Transformer's capability in handling long-range dependencies. RepLKNet \cite{Ding2022} adopts 31$\times$31 large convolution kernels to expand the receptive field, while mitigating the drastic increase in computational load through DWConv. SLaK \cite{Liu2023} further pushes the convolution kernel size to 51 by introducing dynamic sparsity into matrix kernels. OKNet \cite{Cui2024} captures multi-scale receptive fields and modulates large-scale information using 63$\times$63 large convolution kernels with varied shapes. PeLK \cite{Chen2024} proposes peripheral convolution that simulates human visual perception and effectively reduces convolutional parameters through parameter sharing, allowing the convolution kernel size to reach an astonishing 101.
\section{Proposed Method}
\begin{figure*}[t]
\centering
\includegraphics[width=1\textwidth]{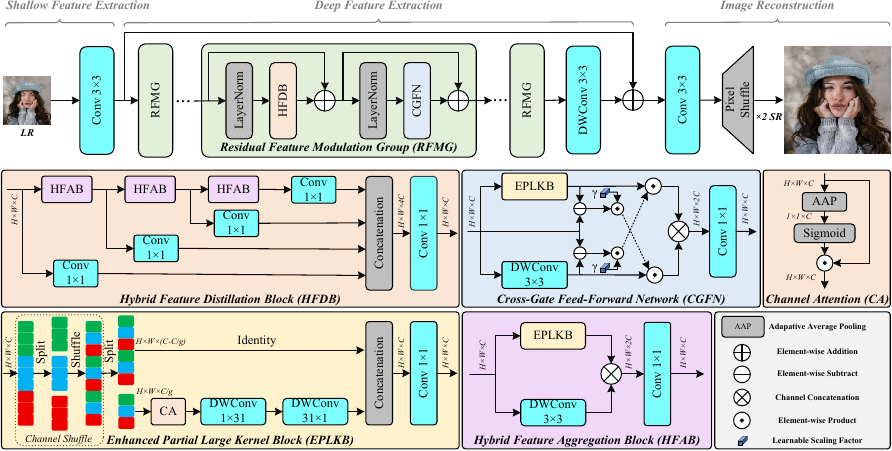}
\caption{The overall architecture of Large Kernel Modulation Network (LKMN), and the detail structure of Hybrid Feature Distillation Block (HFDB) and Cross-Gate Feed Forward Network (CGFN).}
\label{fig2}
\end{figure*}

This section elaborates on the structure of each component in the proposed model. Firstly, Fig. \ref{fig2} illustrates the overall architecture of LKMN, which can be divided into three parts based on their functions: a 3$\times$3 convolutional layer for shallow feature extraction, multiple stacked Residual Feature Modulation Groups (RFMGs) for deep feature extraction, and a sub-pixel layer \cite{Shi2016} for image reconstruction. Next, we will detail the core components within the RFMG.
\subsection{Enhanced Partial Large Kernel Block}
To minimize the model complexity to the greatest extent possible, Following PLKSR \cite{Lee2024} and FasterNet \cite{Chen2023}, we also perform feature extraction only on partial channels. The key difference lies in that we further decompose the convolution kernel into strip convolutions, which enables a further increase in the size of the convolution kernel. Considering that directly splitting channels fails to effectively capture feature information from the remaining channels, we introduce a channel shuffle operation \cite{Zhang2018} to enhance information interaction between the selected channel groups and the remaining channels, and adopt channel attention \cite{Hu2018} to further strengthen the focus on channel information. Given the input feature $I_{in}^{C}$ with $C$ channels, the output after channel shuffle and split can be expressed as:
\begin{equation}
I_{in}^{C/g}, I_{in}^{C-C/g} = F_{Split}^{g}(F_{Shuffle}^{g}(I_{in}^{C}))
\end{equation}
where $g$ denotes the group number of channel shuffle and split. Subsequently, we perform feature extraction on the separated partial channels feature $I_{in}^{C/g}$ while keeping the remaining channels feature $I_{in}^{C-C/g}$ unchanged. This process can be expressed as follows:
\begin{equation}
I_{out}^{C/g} = DWConv_{31\times1}(DWConv_{1\times31}(F_{CA}(I_{in}^{C/g})))
\end{equation}
where $F_{CA}$ indicates the channel attention module. Finally, we concatenate the two parts of features along the channel dimension and utilize a 1$\times$1 convolution for information fusion. The final output $I_{out}$ of the EPLKB can be expressed as follows:
\begin{equation}
I_{out} = Conv_{1\times1}(I_{out}^{C/g}\otimes I_{in}^{C-C/g})
\end{equation}
where $\otimes$ indicates the channel concatenation operation.
\subsection{Hybrid Feature Distillation Block}
To simultaneously capture and fuse non-local and local features, we design a Hybrid Feature Aggregation Block (HFAB), which aggregates local features extracted by DWConv3$\times$3 and non-local features extracted by EPLKB through channel concatenation and 1$\times$1 convolution. Given the input $I_{in}$, the output $I_{out}$ of HFAB can be expressed as follows:
\begin{equation}
I_{out} = Conv_{1\times1}(DWConv_{3\times3}(I_{in})\otimes F_{EPLKB}(I_{in}))
\end{equation}
Subsequently, we construct a Hybrid Feature Distillation Block (HFDB) via embedding multiple HFABs into the main branch of the feature distillation structure to enhance the feature extraction capability, while retaining the 1$\times$1 convolution to eliminate redundant information. Given the input $I_{in}$, this process of HFDB can be expressed as follows:
\begin{equation}
\begin{aligned}
I_{m}^{1},I_{d}^{1}&=HFAB_{1}(I_{in}),Conv_{1\times1}^{1}(I_{in})\\
I_{m}^{2},I_{d}^{2}&=HFAB_{2}(I_{m}^{1}),Conv_{1\times1}^{2}(I_{m}^{1})\\
I_{m}^{3},I_{d}^{3}&=HFAB_{3}(I_{m}^{2}),Conv_{1\times1}^{3}(I_{m}^{2})\\
\end{aligned}
\end{equation}
where $I_{m}^{\it{i}}$ and $I_{d}^{\it{i}}$ denote the $i$th layer output of the main branch and distillation branch. Finally, we aggregate the feature information from the distillation branch and the main branch via channel concatenation and 1$\times$1 convolution. The output $I_{out}$ of the HFDB can be expressed as follows:
\begin{equation}
I_{out}=Conv_{1\times1}(I_{d}^{1} \otimes I_{d}^{2} \otimes I_{d}^{3} \otimes I_{m}^{3})\\
\end{equation}
\par
\subsection{Cross-Gate Feed-Forward Network}
To further enhance the model's capability in modeling and capturing feature information, we first compute the discrepancy information between the input features and the non-local features extracted by EPLKB, as well as between the input features and the local features extracted by DWConv3$\times$3. We then utilize a learnable scaling factor $\gamma$ to adaptively adjust and refine the discrepancy information. By matching the discrepancy information from the EPLKB branch with that from the DWConv3$\times$3 branch, a gating mechanism for modulating local features is constructed. Correspondingly, matching the discrepancy information from the DWConv3$\times$3 branch with that from the EPLKB branch enables the construction of a gating mechanism for modulating non-local features. Finally, the feature information from the two branches is concatenated along the channel dimension and fused using a 1$\times$1 convolution. Given the input feature $I_{in}$, the whole process can be expressed as:
\begin{equation}
I_{E},I_{D}=F_{EPLKB}(I_{in}), F_{DWConv3\times3}(I_{in})\\
\end{equation}
\begin{equation}
I_{E}^{sub},I_{D}^{sub}=\gamma \odot (I_{in}-I_{E}), \gamma \odot (I_{in}-I_{D})\\
\end{equation}
\begin{equation}
I_{out}^{E},I_{out}^{D}=I_{E} \odot I_{D}^{sub}, I_{D} \odot I_{E}^{sub}\\
\end{equation}
\begin{equation}
I_{out} = Conv_{1\times1}(I_{out}^{E}\otimes I_{out}^{D})
\end{equation}
where $I_{E}^{sub}$ and $I_{D}^{sub}$ indicate discrepancy information of EPLKB branch and DWConv3$\times$3 branch, respectively. $\gamma$ is the learnable scaling factor. $\otimes$ and $\odot$ are the operation of channel concatenation and element-wise product, respectively. $I_{out}$ is the final output of CGFN.
\begin{table*}[t]
\centering
\caption{Quantitative comparison between the proposed method and other SOTA tiny-version lightweight SR methods on benchmark datasets. The best and second best results are highlighted in \textcolor{red}{red} and \textcolor{blue}{blue} respectively. \#Flops is measured by recovering a 1280$\times$720 HR image. ``+" denotes the self-ensemble strategy \cite{Lim2017} is used in testing.}
\label{tab:Quantitative comparison with tiny SR methods}
\resizebox{\linewidth}{!}{
\begin{tabular}{ccccccccccc}
\toprule
\multirow{2}{*}{Scale} & \multirow{2}{*}{Method} & \multirow{2}{*}{Years} & \multirow{2}{*}{Dataset} & \#Params & \#Flops & Set5 & Set14 & BSD100 & Urban100 & Manga109\\
& & & & [K] & [G] & PSNR / SSIM & PSNR / SSIM & PSNR / SSIM & PSNR / SSIM & PSNR / SSIM\\
\midrule%第二道横线
\multirow{9}{*}{×2}
& PAN & ECCV2020 & DF2K & 261 & 70.6 & 38.00 / 0.9605 & 33.59 / 0.9181 & 32.18 / 0.8997 & 32.01 / 0.9273 & 38.70 / 0.9773\\
& SAFMN & ICCV2023 & DF2K & 228 & 52 & 38.00 / 0.9605 & 33.54 / 0.9177 & 32.16 / 0.8995 & 31.84 / 0.9256 & 38.71 / 0.9771\\
& MAN-tiny & CVPRW2024 & DF2K & 134 & 30 & 37.93 / 0.9604 & 33.48 / 0.9171 & 32.13 / 0.8993 & 31.75 / 0.9250 & 38.58 / 0.9770\\
& SMFANet & ECCV2024 & DF2K & 186 & 41 & 38.08 / 0.9607 & 33.65 / 0.9186 & 32.22 / 0.9002 & 32.20 / 0.9282 & 39.11 / 0.9779\\
& SeemoRe-T & ICML2024 & DF2K & 220 & 46.1 & 38.06 / 0.9608 & 33.65 / 0.9186 & 32.23 / 0.9004 & 32.22 / 0.9286 & 39.01 / 0.9777\\
& CFSR & TIP2024 & DF2K & 291 & 62.6 & 38.07 / 0.9607 & 33.74 / 0.9192 & 32.24 / 0.9005 & 32.28 / 0.9300 & 39.00 / 0.9778\\
& EARFA-light+ & ACMMM2024 & DF2K & 199 & 44.1 & 38.05 / 0.9608 & 33.65 / 0.9188 & 32.23 / 0.9005 & 32.28 / 0.9298 & 39.10 / 0.9781\\
\rowcolor{gray!20}
\cellcolor{white!} & \textbf{LKMN} & Ours & DF2K & 206 & 46 & \textcolor{blue}{\textbf{38.14}} / \textcolor{blue}{\textbf{0.9610}} & \textcolor{blue}{\textbf{33.83}} / \textcolor{blue}{\textbf{0.9198}} & \textcolor{blue}{\textbf{32.31}} / \textcolor{blue}{\textbf{0.9016}} & \textcolor{blue}{\textbf{32.54}} / \textcolor{blue}{\textbf{0.9323}} & \textcolor{blue}{\textbf{39.18}} / \textcolor{blue}{\textbf{0.9782}}\\
\rowcolor{gray!20}
\cellcolor{white!} & \textbf{LKMN+} & Ours & DF2K & 206 & 46 & \textcolor{red}{\textbf{38.17}} / \textcolor{red}{\textbf{0.9611}} & \textcolor{red}{\textbf{33.89}} / \textcolor{red}{\textbf{0.9201}} & \textcolor{red}{\textbf{32.35}} / \textcolor{red}{\textbf{0.9019}} & \textcolor{red}{\textbf{32.70}} / \textcolor{red}{\textbf{0.9333}} & \textcolor{red}{\textbf{39.30}} / \textcolor{red}{\textbf{0.9785}}\\
\midrule%第二道横线
\multirow{9}{*}{×3}
& PAN & ECCV2020 & DF2K & 261 & 39.1 & 34.10 / 0.9271 & 30.36 / 0.8423 & 29.11 / 0.8050 & 28.11 / 0.8511 & 33.61 / 0.9448\\
& SAFMN & ICCV2023 & DF2K & 233 & 23 & 34.34 / 0.9267 & 30.33 / 0.8418 & 29.08 / 0.8048 & 27.95 / 0.8474 & 33.52 / 0.9437\\
& MAN-tiny & CVPRW2024 & DF2K & 141 & 14 & 34.24 / 0.9259 & 30.25 / 0.8405 & 29.04 / 0.8046 & 27.85 / 0.8465 & 33.28 / 0.9426\\
& SMFANet & ECCV2024 & DF2K & 191 & 19 & 34.42 / 0.9274 & 30.41 / 0.8430 & 29.16 / 0.8065 & 28.22 / 0.8523 & 33.96 / 0.9460\\
& SeemoRe-T & ICML2024 & DF2K & 225 & 21 & 34.46 / 0.9276 & 30.44 / 0.8445 & 29.15 / 0.8063 & 28.27 / 0.8538 & 33.92 / 0.9460\\
& CFSR & TIP2024 & DF2K & 298 & 28.5 & 34.50 / 0.9279 & 30.44 / 0.8437 & 29.16 / 0.8066 & 28.29 / 0.8533 & 33.86 / 0.9462\\
& EARFA-light+ & ACMMM2024 & DF2K & 203 & 20 & 34.48 / 0.9280 & 30.44 / 0.8438 & 29.16 / 0.8067 & 28.29 / 0.8549 & 33.94 / 0.9466\\
\rowcolor{gray!20}
\cellcolor{white!}& \textbf{LKMN} & Ours & DF2K & 211 & 21 & \textcolor{blue}{\textbf{34.55}} / \textcolor{blue}{\textbf{0.9284}} & \textcolor{blue}{\textbf{30.51}} / \textcolor{blue}{\textbf{0.8457}} & \textcolor{blue}{\textbf{29.22}} / \textcolor{blue}{\textbf{0.8087}} & \textcolor{blue}{\textbf{28.47}} / \textcolor{blue}{\textbf{0.8582}} & \textcolor{blue}{\textbf{34.07}} / \textcolor{blue}{\textbf{0.9472}}\\
\rowcolor{gray!20}
\cellcolor{white!} & \textbf{LKMN+} & Ours & DF2K & 211 & 21 & \textcolor{red}{\textbf{34.59}} / \textcolor{red}{\textbf{0.9287}} & \textcolor{red}{\textbf{30.56}} / \textcolor{red}{\textbf{0.8464}} & \textcolor{red}{\textbf{29.25}} / \textcolor{red}{\textbf{0.8092}} & \textcolor{red}{\textbf{28.56}} / \textcolor{red}{\textbf{0.8595}} & \textcolor{red}{\textbf{34.24}} / \textcolor{red}{\textbf{0.9480}}\\
\midrule%第二道横线
\multirow{10}{*}{×4}
& PAN & ECCV2020 & DF2K & 272 & 28.2 & 32.13 / 0.8948 & 28.61 / 0.7822 & 27.59 / 0.7363 & 26.11 / 0.7854 & 30.51 / 0.9095\\
& SAFMN & ICCV2023 & DF2K & 240 & 14 & 32.18 / 0.8948 & 28.60 / 0.7813 & 27.58 / 0.7359 & 25.97 / 0.7809 & 30.43 / 0.9063\\
& MAN-Tiny & CVPRW2024 & DF2K & 150 & 8.4 & 32.07 / 0.8930 & 28.53 / 0.7801 & 27.51 / 0.7345 & 25.84 / 0.7786 & 30.18 / 0.9047\\
& SMFANet & ECCV2024 & DF2K & 197 & 11 & 32.25 / 0.8956 & 28.71 / 0.7833 & 27.64 / 0.7377 & 26.18 / 0.7862 & 30.82 / 0.9104\\
& SeemoRe-T & ICML2024 & DF2K & 232 & 11 & 32.31 / 0.8965 & 28.72 / 0.7840 & 27.65 / 0.7384 & 26.23 / 0.7883 & 30.82 / 0.9107\\
& CFSR & TIP2024 & DF2K & 307 & 17.5 & 32.33 / 0.8964 & 28.73 / 0.7842 & 27.63 / 0.7381 & 26.21 / 0.7897 & 30.72 / 0.9111\\
& EARFA-light+ & ACMMM2024 & DF2K & 209 & 11.6 & 32.33 / 0.8964 & 28.68 / 0.7832 & 27.64 / 0.7382 & 26.20 / 0.7889 & 30.75 / 0.9115\\
& SLVR-S & CVPR2025 & DF2K & 158 & 8.5 & 32.20 / 0.8952 & 28.65 / 0.7822 & 27.57 / 0.7361 & 26.08 / 0.7854 & 30.53 / 0.9084\\
\rowcolor{gray!20}
\cellcolor{white!} & \textbf{LKMN} & Ours & DF2K & 218 & 12.2 & \textcolor{blue}{\textbf{32.40}} / \textcolor{blue}{\textbf{0.8976}} & \textcolor{blue}{\textbf{28.78}} / \textcolor{blue}{\textbf{0.7853}} & \textcolor{blue}{\textbf{27.70}} / \textcolor{blue}{\textbf{0.7399}} & \textcolor{blue}{\textbf{26.35}} / \textcolor{blue}{\textbf{0.7918}} & \textcolor{blue}{\textbf{30.97}} / \textcolor{blue}{\textbf{0.9126}}\\
\rowcolor{gray!20}
\cellcolor{white!} & \textbf{LKMN+} & Ours & DF2K & 218 & 12.2 & \textcolor{red}{\textbf{32.47}} / \textcolor{red}{\textbf{0.8983}} & \textcolor{red}{\textbf{28.82}} / \textcolor{red}{\textbf{0.7863}} & \textcolor{red}{\textbf{27.73}} / \textcolor{red}{\textbf{0.7405}} & \textcolor{red}{\textbf{26.44}} / \textcolor{red}{\textbf{0.7938}} & \textcolor{red}{\textbf{31.15}} / \textcolor{red}{\textbf{0.9144}}\\
\bottomrule
\end{tabular}}
\end{table*}
\section{Experimental Setting}
\subsection{Datasets and Indices}
Following previous works \cite{Wang2024,Zheng2024,Zamfir2024}, we adopt DIV2K \cite{Agustsson2017} and Flickr2K \cite{Lim2017} datasets for training. LR images are generated from HR images by Matlab bicubic downsampling. We adopt five datasets-Set5 \cite{Bevilacqua2012}, Set14 \cite{Zeyde2012}, BSD100 \cite{Martin2001}, Urban100 \cite{Huang2015} and Manga109 \cite{Matsui2017} for testing. Meanwhile, we calculate PSNR and SSIM results on the Y-channel in the YCbCr color space.
\subsection{Implement Detail}
The training HR images are decomposed into 480$\times$480 small pieces through sliding window slicing operation for faster training, while with random rotation, horizontal and vertical flipping for data augment. The cropped LR patch size and batch-size are set to 48 and 64, respectively. Following \cite{Zamfir2024,Xie2023}, we optimize $L_{1}$ loss and FFT loss using Adan optimizer for 1000K iterations, especially 500K iterations for LKMN-L when training on DIV2K dataset. The initial and minimum learning rates are set to 5$\times$$10^{-3}$ and 1$\times$$10^{-6}$, which are updated according to the Cosine Annealing scheme. All experiments are performed with the PyTorch framework on an NVIDIA RTX 3090 GPU. The number of RFMGs and channels are set to \{8, 12\} and \{36, 64\} for \{LKMN, LKMN-L\}, respectively. Meanwhile, the shuffle group $g$ and kernel size in EPLKB are both set to 4 and 31, respectively.
\par
We have listed the hyperparameter settings for different LKMN versions to facilitate better understanding and reproduction, as shown in Table \ref{Implementation Details} below.
\begin{table}[!ht]
\centering
\caption{Implementation Details.}
\label{Implementation Details}
\resizebox{\columnwidth}{!}{
\begin{tabular}{cccc}
\toprule%第一道横线
Hyperparameters & LKMN & LKMN-L & LKMN-L\\
\midrule
Number of RFMGs & 8 & \multicolumn{2}{c}{12}\\
Number of Channels & 36 & \multicolumn{2}{c}{64}\\
Large kernel size & \multicolumn{3}{c}{31} \\
Shuffle group & \multicolumn{3}{c}{4} \\
Patch size & \multicolumn{3}{c}{$48\times48$} \\
Batch size & \multicolumn{3}{c}{64} \\
Training datasets & DF2K & DIV2K & DF2K \\
Optimizer & \multicolumn{3}{c}{Adan} \\
LR-Decay strategy & \multicolumn{3}{c}{Cosine Annealing scheme} \\
Total iterations  & $1000k$ & $500k$ & $1000k$ \\
Initial LR & \multicolumn{3}{c}{$5\times10^{-3}$} \\
Minimum LR & \multicolumn{3}{c}{$1\times10^{-6}$} \\
Initial Scaler & \multicolumn{3}{c}{$1\times10^{-5}$} \\
L1 loss weight & \multicolumn{3}{c}{1} \\
FFT loss weight & \multicolumn{3}{c}{0.05} \\
\bottomrule
\end{tabular}}
\end{table}
\subsection{Comparison with state-of-the-art methods}
To evaluate the performance of the proposed LKMN, we compare it with several SOTA lightweight SR models, including PAN \cite{Zhao2020}, SAFMN \cite{Sun2023}, MAN \cite{Wang2024b}, SMFANet \cite{Zheng2024}, SeemoRe \cite{Zamfir2024}, CFSR \cite{Wu2024}, EARFA \cite{Zhao2024}, SLVR-S \cite{Ni2025}, SwinIR-light \cite{Liang2021}, SRFormer-light \cite{Zhou2023}, MambaIR-light \cite{Guo2024}, MaIR-tiny \cite{Li2025}, CAMixerSR \cite{Wang2024a} and DAT-light \cite{Chen2023a}.
\subsubsection{Quantitative comparison}
Table \ref{tab:Quantitative comparison with tiny SR methods} and Table \ref{tab:Quantitative comparison with large SR methods} present the SR performance comparisons results. For LKMN and LKMN-L, both of them almost achieve the best SR performance across all datasets compared with CNN-based and Transformer-based models. Meanwhile, when LKMN-L is training only on the DIV2K dataset for a fair comparison with Transformer-based models, it still achieves superior or comparable SR performance. The experimental results demonstrate the effectiveness of the proposed LKMN.
\begin{table*}[!ht]
\centering
\caption{Quantitative comparison between the proposed method and other SOTA large-version lightweight SR methods on benchmark datasets. The best and second best results are highlighted in \textcolor{red}{red} and \textcolor{blue}{blue} respectively. \#Flops is measured by recovering a 1280$\times$720 HR image. ``+" denotes the self-ensemble strategy \cite{Lim2017} is used in testing.}
\label{tab:Quantitative comparison with large SR methods}
\resizebox{1\linewidth}{!}{
\begin{tabular}{ccccccccccc}
\toprule
\multirow{2}{*}{Scale} & \multirow{2}{*}{Method} & \multirow{2}{*}{Years} & \multirow{2}{*}{Dataset} & \#Params & \#Flops & Set5 & Set14 & BSD100 & Urban100 & Manga109\\
& & & & [K] & [G] & PSNR / SSIM & PSNR / SSIM & PSNR / SSIM & PSNR / SSIM & PSNR / SSIM\\
\midrule%第二道横线
\multirow{11}{*}{×2}
& SwinIR-light & ICCVW2021 & DIV2K & 910 & 244 & 38.14 / 0.9611 & 33.86 / 0.9206 & 32.31 / 0.9012 & 32.76 / 0.9340 & 39.12 / 0.9783\\
& ELAN-light & ECCV2022 & DIV2K & 621 & 203 & 38.17 / 0.9611 & 33.94 / 0.9207 & 32.30 / 0.9012 & 32.76 / 0.9340 & 39.12 / 0.9783\\
& SRFormer-light & ICCV2023 & DIV2K & 853 & 236 & 38.23 / 0.9613 & 33.94 / 0.9209 & 32.36 / 0.9019 & 32.91 / 0.9353 & 39.28 / 0.9785\\
& MambaIR-light & ECCV2024 & DIV2K & 859 & 198.1 & 38.16 / 0.9610 & 34.00 / 0.9212 & 32.34 / 0.9017 & 32.92 / 0.9356 & 39.31 / 0.9779\\
& MaIR-tiny & CVPR2025 & DIV2K & 878 & 207.8 & 38.18 / 0.9610 & 33.89 / 0.9209 & 32.31 / 0.9013 & 32.89 / 0.9346 & 39.22 / 0.9778\\
& \textbf{LKMN-L} & Ours & DIV2K & 889 & 201 & 38.29 / 0.9615 & 34.11 / 0.9218 & 32.36 / 0.9022 & 32.96 / 0.9355 & 39.23 / 0.9783\\
& DAT-light & ICCV2023 & DF2K & 553 & 194.3 & 38.24 / 0.9614 & 34.01 / 0.9214 & 32.34 / 0.9019 & 32.89 / 0.9346 & 39.49 / 0.9788\\
& MAN-light & CVPRW2024 & DF2K & 823 & 184 & 38.20 / 0.9613 & 33.95 / 0.9214 & 32.36 / 0.9022 & 32.92 / 0.9364 & 39.40 / 0.9786\\
& SeemoRe-L & ICML2024 & DF2K & 953 & 202 & 38.27 / 0.9616 & 34.01 / 0.9210 & 34.35 / 0.9018 & 32.87 / 0.9344 & 39.49 / 0.9790\\
\rowcolor{gray!20}
\cellcolor{white!} & \textbf{LKMN-L} & Ours & DF2K & 889 & 201 & \textcolor{blue}{\textbf{38.32}} / \textcolor{blue}{\textbf{0.9618}} & \textcolor{blue}{\textbf{34.20}} / \textcolor{blue}{\textbf{0.9223}} & \textcolor{blue}{\textbf{32.43}} / \textcolor{blue}{\textbf{0.9030}} & \textcolor{blue}{\textbf{33.13}} / \textcolor{blue}{\textbf{0.9377}} & \textcolor{blue}{\textbf{39.54}} / \textcolor{blue}{\textbf{0.9791}}\\
\rowcolor{gray!20}
\cellcolor{white!} & \textbf{LKMN-L+} & Ours & DF2K & 889 & 201 & \textcolor{red}{\textbf{38.35}} / \textcolor{red}{\textbf{0.9620}} & \textcolor{red}{\textbf{34.26}} / \textcolor{red}{\textbf{0.9226}} & \textcolor{red}{\textbf{32.46}} / \textcolor{red}{\textbf{0.9034}} & \textcolor{red}{\textbf{33.32}} / \textcolor{red}{\textbf{0.9390}} & \textcolor{red}{\textbf{39.70}} / \textcolor{red}{\textbf{0.9795}}\\
\midrule%第二道横线
\multirow{11}{*}{×3}
& SwinIR-light & ICCVW2021 & DIV2K & 918 & 114 & 34.62 / 0.9289 & 30.54 / 0.8463 & 29.20 / 0.8082 & 28.66 / 0.8624 & 33.98 / 0.9478\\
& ELAN-light & ECCV2022 & DIV2K & 629 & 90.1 & 34.61 / 0.9288 & 30.55 / 0.8463 & 29.21 / 0.8081 & 28.69 / 0.8624 & 34.00 / 0.9478\\
& SRFormer-light & ICCV2023 & DIV2K & 861 & 105 & 34.67 / 0.9296 & 30.57 / 0.8469 & 29.26 / 0.8099 & 28.81 / 0.8655 & 34.19 / 0.9489\\
& MambaIR-light & ECCV2024 & DIV2K & 867 & 88.7 & 34.72 / 0.9296 & 30.63 / 0.8475 & 29.29 / 0.8099 & 29.00 / 0.8689 & 34.39 / 0.9495\\
& MaIR-tiny & CVPR2025 & DIV2K & 886 & 93 & 34.68 / 0.9292 & 30.54 / 0.8461 & 29.25 / 0.8088 & 28.83 / 0.8651 & 34.21 / 0.9484\\
& \textbf{LKMN-L} & Ours & DIV2K & 897 & 90.1 & 34.71 / 0.9298 & 30.63 / 0.8479 & 29.29 / 0.8107 & 28.87 / 0.8661 & 34.32 / 0.9492\\
& DAT-light & ICCV2023 & DF2K & 561 & 88.6 & \textcolor{blue}{34.76} / 0.9299 & 30.63 / 0.8474 & 29.29 / 0.8103 & 28.89 / 0.8666 & 34.55 / 0.9501\\
& MAN-light & CVPRW2024 & DF2K & 832 & 82.7 & 34.66 / 0.9293 & 30.60 / 0.8478 & 29.29 / 0.8105 & 28.87 / 0.8671 & 34.36 / 0.9492\\
& SeemoRe-L & ICML2024 & DF2K & 959 & 90.5 & 34.72 / 0.9297 & 30.60 / 0.8469 & 29.29 / 0.8101 & 28.86 / 0.8653 & 34.53 / 0.9496\\
\rowcolor{gray!20}
\cellcolor{white!} & \textbf{LKMN-L} & Ours & DF2K & 897 & 90.1 & 34.75 / \textcolor{blue}{\textbf{0.9299}} & \textcolor{blue}{\textbf{30.70}} / \textcolor{blue}{\textbf{0.8490}} & \textcolor{blue}{\textbf{29.34}} / \textcolor{blue}{\textbf{0.8118}} & \textcolor{blue}{\textbf{29.07}} / \textcolor{blue}{\textbf{0.8698}} & \textcolor{blue}{\textbf{34.66}} / \textcolor{blue}{\textbf{0.9505}}\\
\rowcolor{gray!20}
\cellcolor{white!} & \textbf{LKMN-L+} & Ours & DF2K & 897 & 90.1 & \textcolor{red}{\textbf{34.80}} / \textcolor{red}{\textbf{0.9302}} & \textcolor{red}{\textbf{30.75}} / \textcolor{red}{\textbf{0.8497}} & \textcolor{red}{\textbf{29.36}} / \textcolor{red}{\textbf{0.8123}} & \textcolor{red}{\textbf{29.18}} / \textcolor{red}{\textbf{0.8713}} & \textcolor{red}{\textbf{34.81}} / \textcolor{red}{\textbf{0.9511}}\\
\midrule%第二道横线
\multirow{12}{*}{×4}
& SwinIR-light & ICCVW2021 & DIV2K & 930 & 65 & 32.44 / 0.8976 & 28.77 / 0.7858 & 27.69 / 0.7406 & 26.47 / 0.7980 & 30.92 / 0.9151\\
& ELAN-light & ECCV2022 & DIV2K & 640 & 54.1 & 32.43 / 0.8975 & 28.78 / 0.7858 & 27.69 / 0.7406 & 26.54 / 0.7982 & 30.92 / 0.9150\\
& SRFormer-light & ICCV2023 & DIV2K & 873 & 63 & 32.51 / 0.8988 & 28.82 / 0.7872 & 27.73 / 0.7422 & 26.67 / 0.8032 & 31.17 / 0.9165\\
& MambaIR-light & ECCV2024 & DIV2K & 879 & 50.6 & 32.51 / 0.8993 & 28.85 / 0.7876 & 27.75 / 0.7423 & 26.75 / 0.8051 & 31.26 / 0.9175\\
& CAMixerSR & CVPR2024 & DIV2K & 765 & 77.9 & 32.51 / 0.8988 & 28.82 / 0.7870 & 27.72 / 0.7416 & 26.63 / 0.8012 & 31.18 / 0.9166\\
& MaIR-Tiny & CVPR2025 & DIV2K & 897 & 53.1 & 32.48 / 0.8985 & 28.81 / 0.7864 & 27.71 / 0.7414 & 26.60 / 0.8013 & 31.13 / 0.9161\\
& \textbf{LKMN-L} & Ours & DIV2K & 909 & 51.4 & 32.56/0.8994 & 28.88/0.7880 & 27.77/0.7425 & 26.65/0.8018 & 31.13/0.9162\\
& DAT-light & ICCV2023& DF2K & 573 & 49.7 & 32.57 / 0.8991 & 28.87 / 0.7879 & 27.79 / 0.7428 & 26.64 / 0.8033 & 31.37 / 0.9178\\
& MAN-light & CVPRW2024 & DF2K & 840 & 47.1 & 32.50 / 0.8988 & 28.87 / 0.7885 & 27.77 / 0.7429 & 26.70 / 0.8052 & 31.25 / 0.9170\\
& SeemoRe-L & ICML2024 & DF2K & 969 & 51.4 & 32.51 / 0.8990 & 28.92 / 0.7888 & 27.78 / 0.7428 & 26.79 / 0.8046 & 31.48 / 0.9181\\
\rowcolor{gray!20}
\cellcolor{white!} & \textbf{LKMN-L} & Ours & DF2K & 909 & 51.4 & \textcolor{blue}{\textbf{32.67}} / \textcolor{blue}{\textbf{0.9005}} & \textcolor{blue}{\textbf{28.96}} / \textcolor{blue}{\textbf{0.7897}} & \textcolor{blue}{\textbf{27.83}} / \textcolor{blue}{\textbf{0.7440}} & \textcolor{blue}{\textbf{26.89}} / \textcolor{blue}{\textbf{0.8075}} & \textcolor{blue}{\textbf{31.60}} / \textcolor{blue}{\textbf{0.9194}}\\
\rowcolor{gray!20}
\cellcolor{white!} & \textbf{LKMN-L+} & Ours & DF2K & 909 & 51.4 & \textcolor{red}{\textbf{32.71}} / \textcolor{red}{\textbf{0.9010}} & \textcolor{red}{\textbf{29.00}} / \textcolor{red}{\textbf{0.7905}} & \textcolor{red}{\textbf{27.85}} / \textcolor{red}{\textbf{0.7446}} & \textcolor{red}{\textbf{27.00}} / \textcolor{red}{\textbf{0.8098}} & \textcolor{red}{\textbf{31.76}} / \textcolor{red}{\textbf{0.9207}}\\
\bottomrule
\end{tabular}}
\end{table*}
\subsubsection{Qualitative comparison}
\begin{figure*}[t]
\centering
\includegraphics[width=1\textwidth]{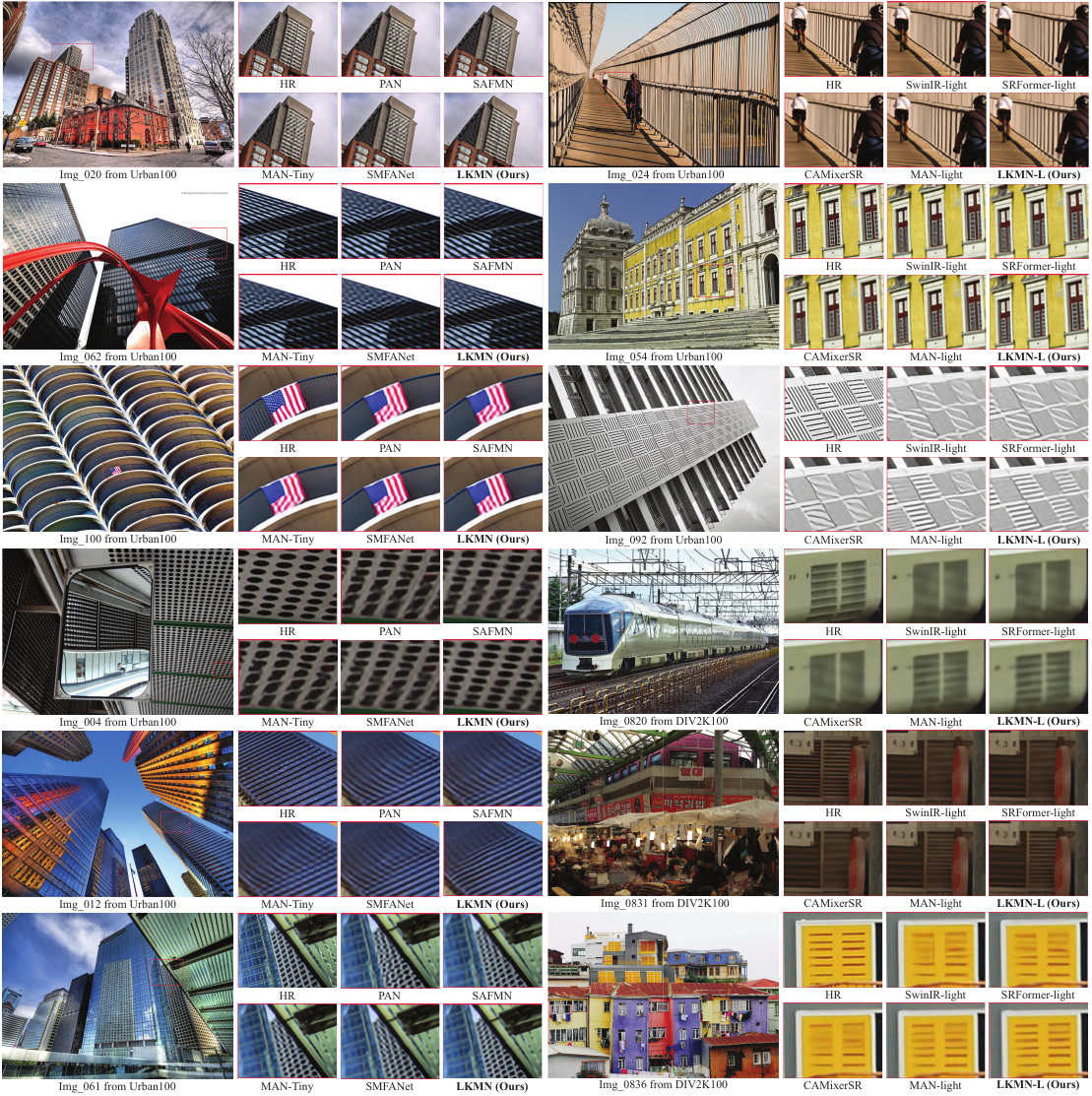}
\caption{Visual comparison with SOTA lightweight SR methods at $\times$4 upscale. Zoom in for best view.}
\label{fig3}
\end{figure*}

Fig. \ref{fig3} presents the visual comparisons between the proposed LKMN and other lightweight SR methods. It can be observed that both the tiny and large versions of our method exhibit stronger image restoration capabilities, being able to reconstruct more image details (e.g., img-062 and img-092) while having less distortion (e.g., img-054 and img-100). The above results also highlight the effectiveness of the proposed LKMN.
\par
\begin{figure}[!ht]
\centering
\includegraphics[width=1\columnwidth]{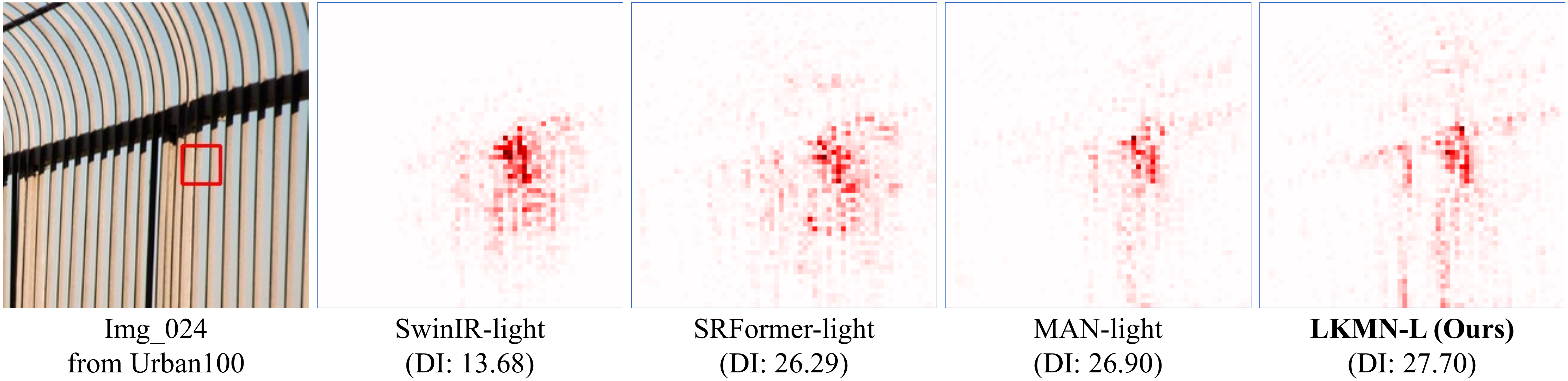}
\caption{LAM comparison on LKMN-L at $\times$4 upscale.}
\label{fig4}
\end{figure}

Meanwhile, the local attribution map (LAM) \cite{Gu2021} is also utilized to analyze the range of receptive field, as shown in Fig. \ref{fig4}. The proposed LKMN-L can obtain higher Diffusion Index (DI) value and larger receptive field, which proves the effectiveness of large kernel convolution.
\subsubsection{GPU memory and latency comparison}
Table \ref{tab: GPU and inference time comparison} shows the GPU memory consumption and inference speed comparison. Although LKMN achieves the optimal SR performance, its inference speed is slightly slower than that of other CNN-based models. LKMN-L achieves the best SR performance and fastest inference speed, which is almost $\times$4.8 times faster than the Transformer-based model DAT-light while with 71.6\% less GPU memory consumption. Meanwhile, LKMN-L is 16\% faster than the CNN-based model MAN-light. The above analysis indicates that the proposed LKMN family achieves a better balance between SR performance and inference speed.
\begin{table*}[t]
\caption{GPU memory consumption and average inference time comparison with SOTA lightweight SR methods, which are all calculated across 100 samples on an NVIDIA RTX 3090 GPU. PSNR results are all calculated on Manga109 dataset.}
\label{tab: GPU and inference time comparison}
\centering
\resizebox{\linewidth}{!}{
\begin{tabular}{cccccccccc}
\toprule
LR input & Upscale & Method & \#GPU Mem. & \#Avg. Time & PSNR & Method & \#GPU Mem. & \#Avg. Time & PSNR\\
\midrule%第二道横线
\multirow{7}{*}{[512,512]}& \multirow{7}{*}{$\times$4} & PAN & 1196.09 M & 53.44 ms & 30.51 & SwinIR-light & 1451.48 M & 1000.89 ms & 30.92\\
& & SAFMN & \textbf{291.96 M} & \textbf{34.78 ms} & 30.43 & SRFormer-light & 1303.83 M & 998.50 ms & 31.17\\
& & SeemoRe-T & 365.02 M & 70.93 ms & 30.82 & DAT-light & 2619.48 M & 1256.98 ms & 31.37\\
& & SMFANet & 365.89 M & 35.51 ms & 30.82 & CAMixerSR & 1489.17 M & 320.41 ms & 31.18\\
& & CFSR & 580.22 M & 88.84 ms & 30.72 & SeemoRe-L & \textbf{470.94 M} & 236.79 ms & 31.48\\
& & MAN-Tiny & 467.64 M & 47.70 ms & 30.18 & MAN-light & 546.50 M & 259.79 ms & 31.25\\
& & LKMN (Ours) & 436.00 M & 84.11 ms & \textbf{30.97} & LKMN-L (Ours) & 774.60 M & \textbf{218.08 ms} & \textbf{31.60}\\
\bottomrule
\end{tabular}}
\end{table*}
\subsubsection{Quantitative visual perception comparison}
To conduct a more comprehensive analysis on the performance of the proposed LKMN, we further employ a subjective evaluation metric LPIPS  to assess the visual perception performance, with the results presented in Table \ref{LPIPS}. For the tiny version, the proposed LKMN achieves the optimal perceptual metrics, demonstrating its superiority over other CNN-based lightweight SR models. For the large version, the proposed LKMN-L yields slightly inferior performance compared to the Transformer-based model CAMixerSR but is marginally better than SwinIR-light, which indicates its potential to outperform Transformer-based lightweight SR models.
\begin{table}[!ht]
\centering
\caption{Perceptual metric (LPIPS$\downarrow$) comparison at $\times$4 upscale.}
\label{LPIPS}
\resizebox{\columnwidth}{!}{
\begin{tabular}{ccccc}
\toprule%第一道横线
Method & Dataset & BSD100 & Urban100 & Manga109\\
\midrule
SAFMN & DF2K & 0.3825 & 0.2488 & 0.1116\\
SMFANet & DF2K & \underline{0.3798} & \underline{0.2423} & \underline{0.1096}\\
\textbf{LKMN} & DF2K & \textbf{0.3766} & \textbf{0.2383} & \textbf{0.1084}\\
\hdashline
SwinIR-light & DIV2K & \underline{0.3672} & 0.2162 & 0.1021\\
CAMixerSR & DIV2K & \textbf{0.3631} & \textbf{0.2150} & \textbf{0.1009}\\
\textbf{LKMN-L} & DIV2K & 0.3673 & \underline{0.2157} & \underline{0.1017}\\
\bottomrule
\end{tabular}}
\end{table}
\subsubsection{Real-world image super-resolution}
\begin{figure*}[!ht]
\centering
\includegraphics[width=1\textwidth]{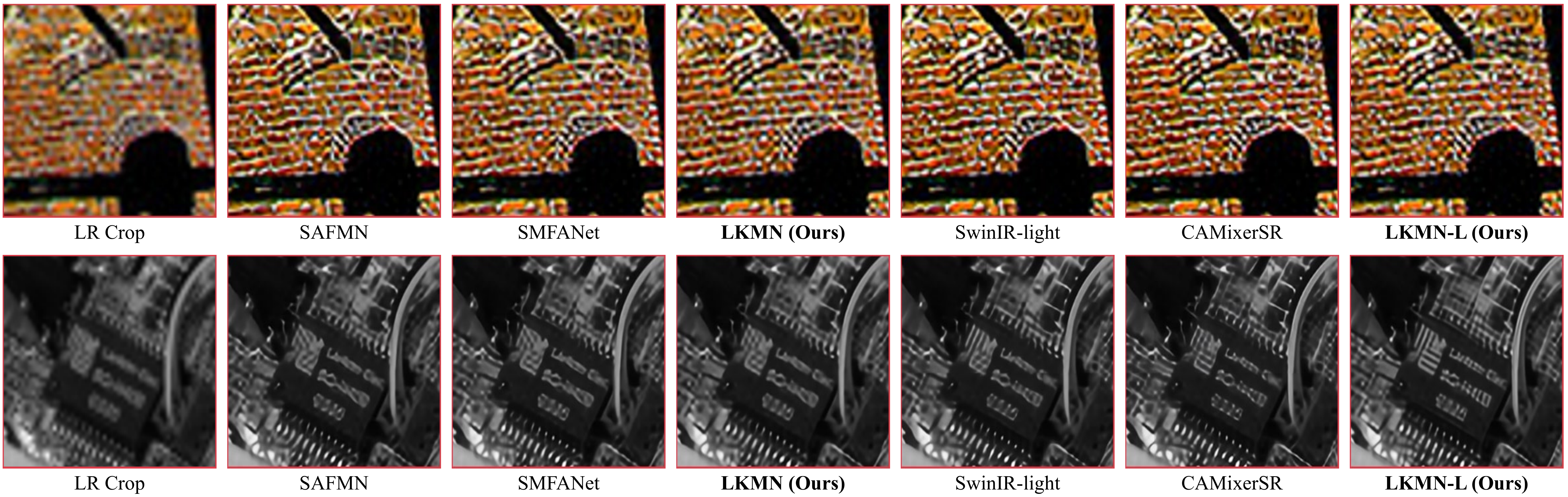}
\caption{Visual comparison between LKMN family with SAFMN \cite{Sun2023}, SMFANet \cite{Zheng2024}, SwinIR-light \cite{Liang2021}, and CAMixerSR \cite{Wang2024a} on \textbf{real-world images} at $\times$4 upscale.}
\label{app_fig1}
\end{figure*}

We also explore the reconstruction capability of the model on real-world images to examine its generalization and practicality, with the results shown in the Fig. \ref{app_fig1}. When not trained on real-world image datasets, the recovery performance of commonly used lightweight SR models is relatively limited. However, it can still be observed that the proposed LKMN-L is able to reconstruct clearer texture details compared to other models, demonstrating the superiority of the proposed model.
\section{Analysis and discussion}
In this section, we conduct all ablation experiments using LKMN and validate on the Manga109 dataset at $\times$4 upscale, with the experimental settings unchanged. Table \ref{Ablation study of block} shows the impact of removing HFDB and CGFN on model performance, demonstrating the effectiveness of the two modules.
\begin{table}[!ht]
\centering
\caption{Ablation study of HFDB and CGFN.}
\label{Ablation study of block}
\resizebox{\columnwidth}{!}{
\begin{tabular}{cccccc}
\toprule%第一道横线
HFDB & CGFN & \#Params & \#Flops & \#Time & PSNR\\
\midrule
$\surd$ & $\times$ & 177K & 10.0G & 58.36ms & 30.85\\
$\times$ & $\surd$ & 57.3K & 3.17G & 27.46ms & 29.57\\
$\surd$ & $\surd$ & 218K & 12.2G & 84.11ms & 30.97\\
\bottomrule
\end{tabular}}
\end{table}
\subsection{Ablation experiments on EPLKB}
We first validate the channel shuffle and channel attention in EPLKB, as shown in the Table \ref{Ablation study of EPLKB}. The model performance decrease when each of these designs are removed, demonstrating their effectiveness. Additionally, when we utilize the full convolution kernel of DWConv31×31, the model parameters, computations and inference time increase by $\times$6.3 times, $\times$6.5 times and 74.4\%, respectively, while the performance decrease instead. This confirms the effectiveness of the strip large convolution kernel design.
\begin{table}[!ht]
\centering
\caption{Ablation study of EPLKB}
\label{Ablation study of EPLKB}
\resizebox{\columnwidth}{!}{
\begin{tabular}{ccccc}
\toprule%第一道横线
Setting & \#Params & \#Flops & \#Time & PSNR\\
\midrule
Base & 218K & 12.2G & 84.11ms & 30.97\\
w/o channel shuffle & 218K & 12.2G & 81.08ms & 30.91\\
w/o channel attention & 218K & 12.2G & 79.84ms & 30.86\\
w DWConv31$\times$31 & 1599K & 91.2G & 146.68ms & 30.93\\
\bottomrule
\end{tabular}}
\end{table}
\par
\begin{figure}[!ht]
\centering
\includegraphics[width=1\columnwidth]{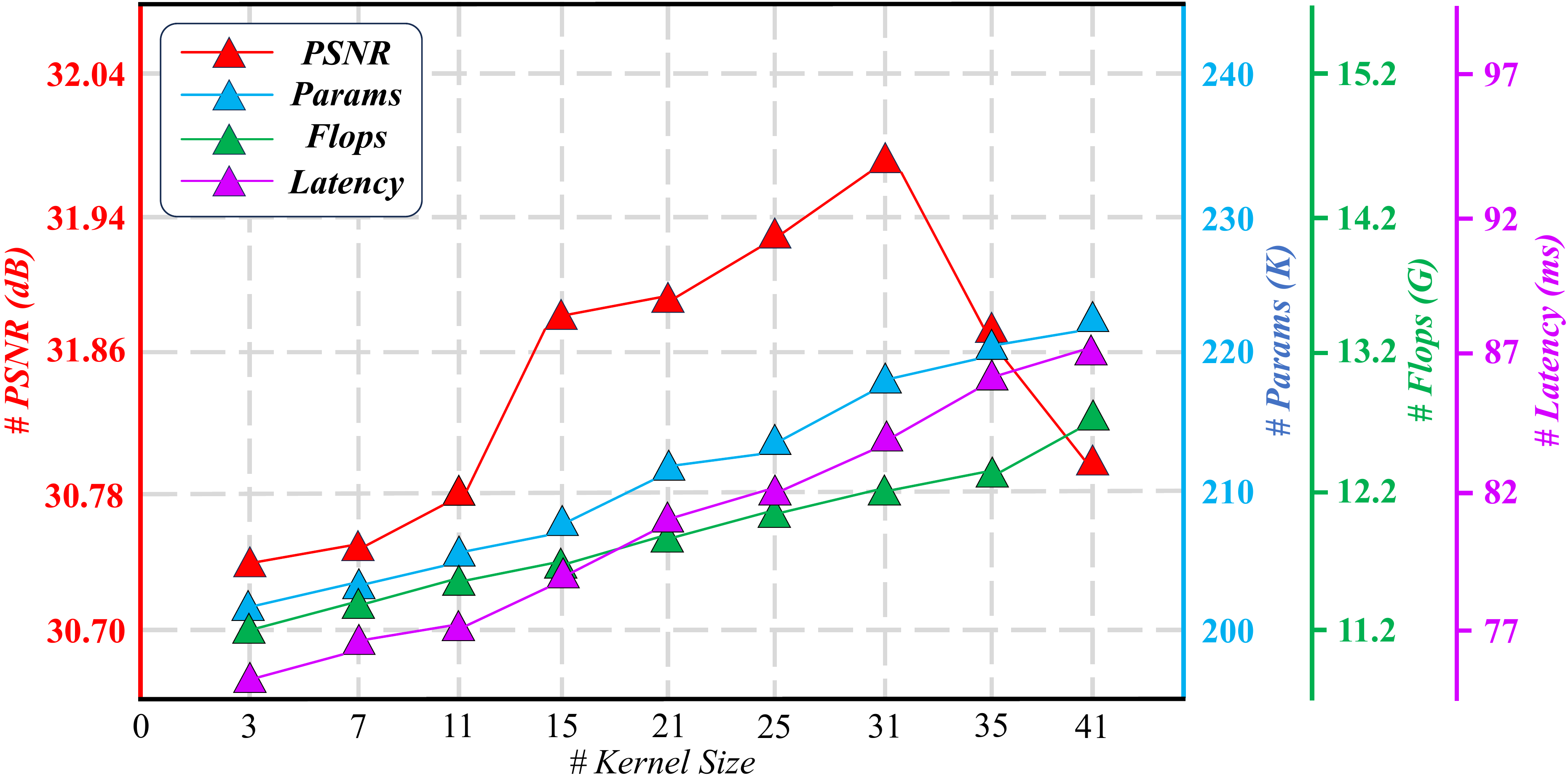}
\caption{Analysis on kernel size in EPLKB.}
\label{fig5}
\end{figure}
\begin{figure}[!ht]
\centering
\includegraphics[width=1\columnwidth]{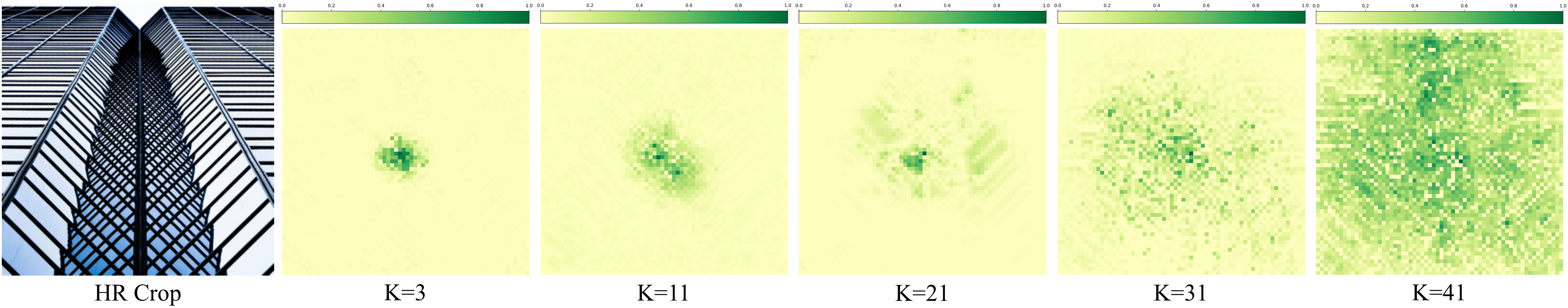}
\caption{Analysis on ERF of kernel size in EPLKB.}
\label{fig6}
\end{figure}

Meanwhile, we analyze the kernel sizes in EPLKB, as illustrated in Fig. \ref{fig5}. As the kernel size increases, the SR performance gradually improves and reaches the optimal when the kernel size is 31. Benefiting from the partial channel strip convolution design, the model complexity and inference time only increase slightly. The performance degradation of the model when the convolution kernel size exceeds 31 may be due to the introduction of more redundant information by excessively large convolution kernels, which remains to be investigated in future work. In addition, we also perform analysis on kernel size using the Effective Receptive Field (ERF) \cite{Ding2022}, as shown in Fig. \ref{fig6}. It can be observed that as the convolution kernel size increases, the range of receptive field also gradually expands. The above experiments all demonstrate the effectiveness of large kernel convolution.
\par
\par
Furthermore, we analyze the impact of the shuffle group $g$ on model performance, as shown in Table \ref{Ablation study of shuffle group}. As the value of $g$ decreases, the number of channels involved in feature extraction gradually increases, leading to a corresponding improvement in SR performance. However, the model complexity and inference time also increase gradually. Considering the design goal of efficient SR models, we ultimately set $g$ to 4 to maintain a better balance among model complexity, SR performance and inference speed.
\begin{table}[!ht]
\centering
\caption{Ablation study of shuffle group in EPLKB}
\label{Ablation study of shuffle group}
\resizebox{\columnwidth}{!}{
\begin{tabular}{ccccc}
\toprule%第一道横线
Shuffle group & \#Params & \#Flops & \#Time & PSNR\\
\midrule
1 & 273K & 15.3G & 121.15ms & 31.02\\
2 & 236K & 13.2G & 97.30ms & 31.00\\
4 & 218K & 12.2G & 84.11ms & 30.97\\
6 & 212K & 11.8G & 79.28ms & 30.84\\
\bottomrule
\end{tabular}}
\end{table}
\subsection{Ablation experiments on CGFN}
Table \ref{Ablation study of CGFN} shows the impact of learnable scaling factor $\gamma$ and cross-gate strategy. The SR performance degrades when the learnable scaling factor $\gamma$ is removed and the cross-gate strategy is replaced with direct gating. In addition, neither of them has a substantial impact on model complexity or inference speed, which all demonstrates their effectiveness.
\begin{table}[!ht]
\centering
\caption{Ablation study of CGFN.}
\label{Ablation study of CGFN}
\resizebox{\columnwidth}{!}{
\begin{tabular}{ccccc}
\toprule%第一道横线
Setting & \#Params & \#Flops & \#Time & PSNR\\
\midrule
Base & 218K & 12.2G & 84.11ms & 30.97\\
w/o scaler $\gamma$ & 217K & 12.2G & 82.51ms & 30.90\\
w/o cross-gate & 218K & 12.2G & 84.11ms & 30.95\\
\bottomrule
\end{tabular}}
\end{table}
\par
\begin{figure}[!ht]
\centering
\includegraphics[width=1\columnwidth]{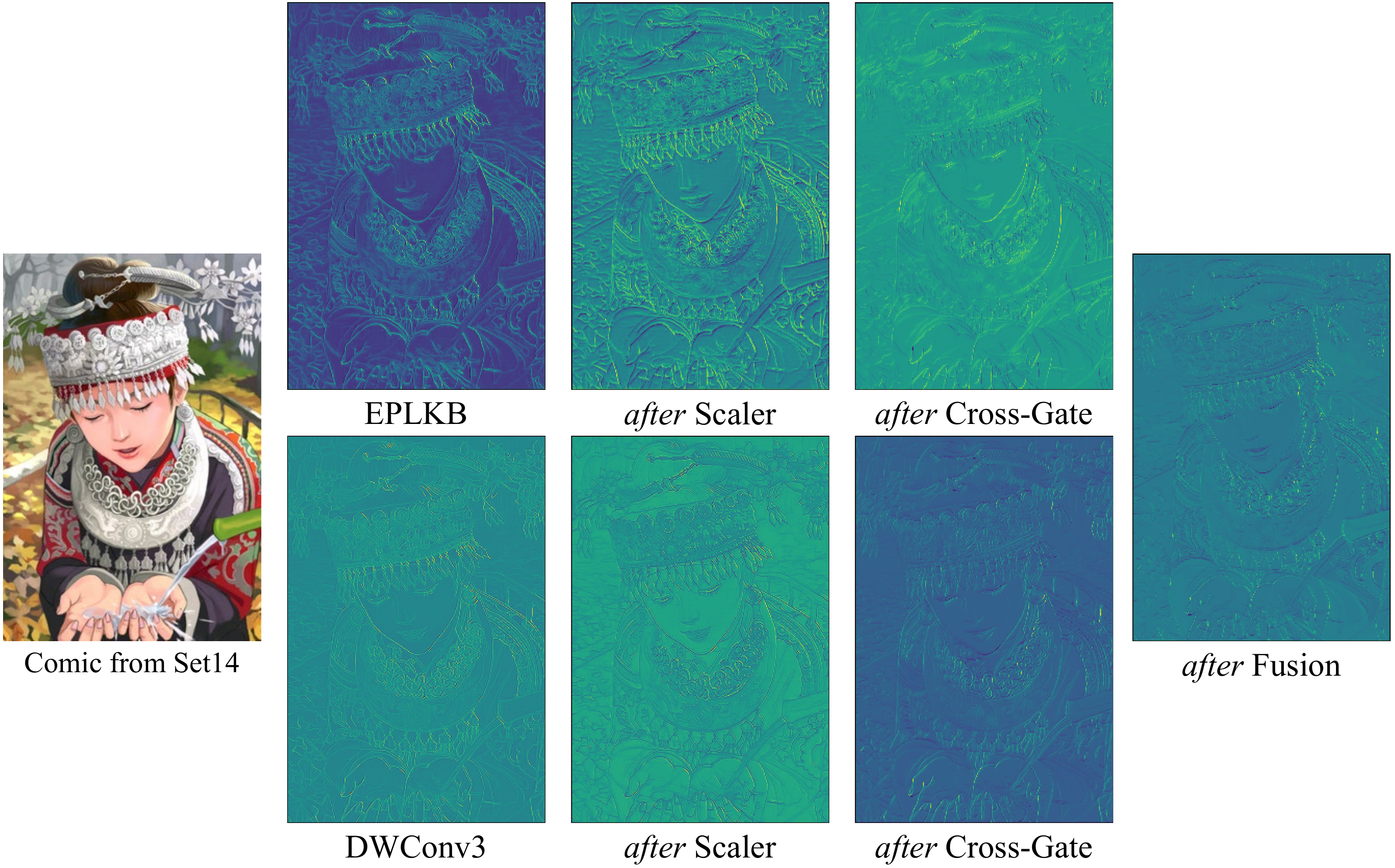}
\caption{Visualization of feature maps in CGFN.}
\label{fig7}
\end{figure}

Furthermore, we analyze the feature map outputs in CGFN, as shown in the Fig. \ref{fig7}. It can be seen that EPLKB tends to extract low-frequency information such as global contours, while DWConv3×3 is responsible for extracting local texture information. Meanwhile, more comprehensive feature information can be obtained by modulating the feature differences, non-local and local features can be modeled more prominently under the effect of cross-gate strategy. The above analysis also proves the effectiveness of the CGFN module.
\subsection{Ablation experiments on loss function}
We analyze the impact of different loss functions on model performance, and the results are shown in Table \ref{loss function}. The FFT loss is used to minimize feature differences in the frequency domain, which can better guide model training from the perspective of low-frequency and high-frequency features. Therefore, compared with the commonly used L1 loss, the model performance is further improved after adding the FFT loss, and this phenomenon is also observed in SeemoRe \cite{Zamfir2024}.
\begin{table}[!ht]
\centering
\caption{The effect of different loss function on SR performance (PSNR$\uparrow$) at $\times$4 upscale.}
\label{loss function}
\resizebox{\columnwidth}{!}{
\begin{tabular}{ccccc}
\toprule%第一道横线
Loss & BSD100 & Urban100 & Manga109 & DIV2K100\\
\midrule
L1 only & 27.67 & 26.34 & 30.78 & 30.59\\
L1+FFT & 27.70 & 26.35 & 30.97 & 30.64\\
\bottomrule
\end{tabular}}
\end{table}
\subsection{Limitations and future work}
The model proposed in this paper has achieved excellent performance, but there are still some limitations. Firstly, although the performance of the proposed model is superior to other SR models, the inference speed of LKMN is slightly slower than that of CNN-based SR models. On the other hand, the optimal performance is obtained when the size of the large convolution kernel used is 31, but further increasing the size leads to a decrease in performance despite expanding the receptive field. The reason for this phenomenon is currently only speculated to be the introduction of noise, and a true explanation has not yet been provided. In future work, we will conduct a comprehensive analysis of the model and explore a more efficient network architecture to further reduce the inference speed. Meanwhile, we will also attempt to use methods such as frequency domain analysis and feature map visualization to further verify the impact of convolution kernel size on model performance.
\section{Conclusion}
In this paper, we propose the Large Kernel Modulation Network (LKMN), a CNN-based efficient SR model. Through the design of partial channel strip large kernel convolutions, it effectively controls the increase in model complexity caused by enlarging kernel size while acquiring strong non-local feature extraction capability. Additionally, by dynamically adjusting the discrepancies between local and non-local features and introducing a cross-gate mechanism, it enhances the specificity of feature fusion, providing a new approach for modeling the complementarity of local and non-local features. These designs enable LKMN to achieve superior SR performance over other lightweight SR methods while retaining the low-latency advantage of CNNs.

\bibliographystyle{IEEEtran}
\bibliography{References}

\begin{thebibliography}{10}
\providecommand{\url}[1]{#1}
\csname url@samestyle\endcsname
\providecommand{\newblock}{\relax}
\providecommand{\bibinfo}[2]{#2}
\providecommand{\BIBentrySTDinterwordspacing}{\spaceskip=0pt\relax}
\providecommand{\BIBentryALTinterwordstretchfactor}{4}
\providecommand{\BIBentryALTinterwordspacing}{\spaceskip=\fontdimen2\font plus
\BIBentryALTinterwordstretchfactor\fontdimen3\font minus
  \fontdimen4\font\relax}
\providecommand{\BIBforeignlanguage}[2]{{%
\expandafter\ifx\csname l@#1\endcsname\relax
\typeout{** WARNING: IEEEtran.bst: No hyphenation pattern has been}%
\typeout{** loaded for the language `#1'. Using the pattern for}%
\typeout{** the default language instead.}%
\else
\language=\csname l@#1\endcsname
\fi
#2}}
\providecommand{\BIBdecl}{\relax}
\BIBdecl

\bibitem{He2016}
K.~He, X.~Zhang, S.~Ren, and J.~Sun, ``Deep residual learning for image
  recognition,'' in \emph{2016 IEEE Conference on Computer Vision and Pattern
  Recognition (CVPR)}, 2016, pp. 770--778.

\bibitem{Liang2021}
J.~Liang, J.~Cao, G.~Sun, K.~Zhang, L.~Van~Gool, and R.~Timofte, ``Swinir:
  Image restoration using swin transformer,'' in \emph{Proceedings of the
  IEEE/CVF international conference on computer vision}, 2021, pp. 1833--1844.

\bibitem{Ren2024}
B.~Ren, Y.~Li, N.~Mehta, R.~Timofte, H.~Yu, C.~Wan, Y.~Hong, B.~Han, Z.~Wu,
  Y.~Zou \emph{et~al.}, ``The ninth ntire 2024 efficient super-resolution
  challenge report,'' in \emph{Proceedings of the IEEE/CVF Conference on
  Computer Vision and Pattern Recognition}, 2024, pp. 6595--6631.

\bibitem{Zhou2023}
Y.~Zhou, Z.~Li, C.-L. Guo, S.~Bai, M.-M. Cheng, and Q.~Hou, ``Srformer:
  Permuted self-attention for single image super-resolution,'' in
  \emph{Proceedings of the IEEE/CVF International Conference on Computer
  Vision}, 2023, pp. 12\,780--12\,791.

\bibitem{Dosovitskiy2021}
A.~Dosovitskiy, L.~Beyer, A.~Kolesnikov, D.~Weissenborn, X.~Zhai,
  T.~Unterthiner, M.~Dehghani, M.~Minderer, G.~Heigold, S.~Gelly, J.~Uszkoreit,
  and N.~Houlsby, ``An image is worth 16x16 words: Transformers for image
  recognition at scale,'' \emph{ICLR}, 2021.

\bibitem{Liu2020}
J.~Liu, J.~Tang, and G.~Wu, ``Residual feature distillation network for
  lightweight image super-resolution,'' in \emph{Computer vision--ECCV 2020
  workshops: Glasgow, UK, August 23--28, 2020, proceedings, part III 16}.\hskip
  1em plus 0.5em minus 0.4em\relax Springer, 2020, pp. 41--55.

\bibitem{Li2022}
Z.~Li, Y.~Liu, X.~Chen, H.~Cai, J.~Gu, Y.~Qiao, and C.~Dong, ``Blueprint
  separable residual network for efficient image super-resolution,'' in
  \emph{Proceedings of the IEEE/CVF conference on computer vision and pattern
  recognition}, 2022, pp. 833--843.

\bibitem{Zhou2022}
L.~Zhou, H.~Cai, J.~Gu, Z.~Li, Y.~Liu, X.~Chen, Y.~Qiao, and C.~Dong,
  ``Efficient image super-resolution using vast-receptive-field attention,'' in
  \emph{European conference on computer vision}.\hskip 1em plus 0.5em minus
  0.4em\relax Springer, 2022, pp. 256--272.

\bibitem{Xie2023}
C.~Xie, X.~Zhang, L.~Li, H.~Meng, T.~Zhang, T.~Li, and X.~Zhao, ``Large kernel
  distillation network for efficient single image super-resolution,'' in
  \emph{Proceedings of the IEEE/CVF Conference on Computer Vision and Pattern
  Recognition}, 2023, pp. 1283--1292.

\bibitem{Lee2024}
D.~Lee, S.~Yun, and Y.~Ro, ``Partial large kernel cnns for efficient
  super-resolution,'' \emph{arXiv preprint arXiv:2404.11848}, 2024.

\bibitem{Dong2014}
C.~Dong, C.~C. Loy, K.~He, and X.~Tang, ``Learning a deep convolutional network
  for image super-resolution,'' in \emph{Computer Vision--ECCV 2014: 13th
  European Conference, Zurich, Switzerland, September 6-12, 2014, Proceedings,
  Part IV 13}.\hskip 1em plus 0.5em minus 0.4em\relax Springer, 2014, pp.
  184--199.

\bibitem{Hui2019}
Z.~Hui, X.~Gao, Y.~Yang, and X.~Wang, ``Lightweight image super-resolution with
  information multi-distillation network,'' in \emph{Proceedings of the 27th
  acm international conference on multimedia}, 2019, pp. 2024--2032.

\bibitem{Zheng2024}
M.~Zheng, L.~Sun, J.~Dong, and J.~Pan, ``Smfanet: A lightweight self-modulation
  feature aggregation network for efficient image super-resolution,'' in
  \emph{ECCV}, 2024.

\bibitem{Zhao2024}
X.~Zhao, L.~Li, C.~Xie, X.~Zhang, T.~Jiang, W.~Lin, S.~Liu, and T.~Li,
  ``Efficient single image super-resolution with entropy attention and
  receptive field augmentation,'' in \emph{Proceedings of the 32nd ACM
  International Conference on Multimedia}, 2024, pp. 1302--1310.

\bibitem{Wang2024a}
Y.~Wang, Y.~Liu, S.~Zhao, J.~Li, and L.~Zhang, ``Camixersr: Only details need
  more" attention",'' in \emph{Proceedings of the IEEE/CVF Conference on
  Computer Vision and Pattern Recognition}, 2024, pp. 25\,837--25\,846.

\bibitem{Park2025}
K.~Park, J.~W. Soh, and N.~I. Cho, ``Efficient attention-sharing information
  distillation transformer for lightweight single image super-resolution,'' in
  \emph{Proceedings of the AAAI Conference on Artificial Intelligence},
  vol.~39, no.~6, 2025, pp. 6416--6424.

\bibitem{Guo2022}
M.-H. Guo, C.-Z. Lu, Q.~Hou, Z.~Liu, M.-M. Cheng, and S.-M. Hu, ``Segnext:
  Rethinking convolutional attention design for semantic segmentation,''
  \emph{Advances in neural information processing systems}, vol.~35, pp.
  1140--1156, 2022.

\bibitem{Ding2022}
X.~Ding, X.~Zhang, J.~Han, and G.~Ding, ``Scaling up your kernels to 31x31:
  Revisiting large kernel design in cnns,'' in \emph{Proceedings of the
  IEEE/CVF conference on computer vision and pattern recognition}, 2022, pp.
  11\,963--11\,975.

\bibitem{Liu2023}
S.~Liu, T.~Chen, X.~Chen, X.~Chen, Q.~Xiao, B.~Wu, T.~K{\"a}rkk{\"a}inen,
  M.~Pechenizkiy, D.~C. Mocanu, and Z.~Wang, ``More convnets in the 2020s:
  Scaling up kernels beyond 51x51 using sparsity,'' in \emph{The Eleventh
  International Conference on Learning Representations}, 2023.

\bibitem{Cui2024}
Y.~Cui, W.~Ren, and A.~Knoll, ``Omni-kernel network for image restoration,'' in
  \emph{Proceedings of the AAAI conference on artificial intelligence},
  vol.~38, no.~2, 2024, pp. 1426--1434.

\bibitem{Chen2024}
H.~Chen, X.~Chu, Y.~Ren, X.~Zhao, and K.~Huang, ``Pelk: Parameter-efficient
  large kernel convnets with peripheral convolution,'' in \emph{Proceedings of
  the IEEE/CVF Conference on Computer Vision and Pattern Recognition}, 2024,
  pp. 5557--5567.

\bibitem{Shi2016}
W.~Shi, J.~Caballero, F.~Husz{\'a}r, J.~Totz, A.~P. Aitken, R.~Bishop,
  D.~Rueckert, and Z.~Wang, ``Real-time single image and video super-resolution
  using an efficient sub-pixel convolutional neural network,'' in
  \emph{Proceedings of the IEEE conference on computer vision and pattern
  recognition}, 2016, pp. 1874--1883.

\bibitem{Chen2023}
J.~Chen, S.-h. Kao, H.~He, W.~Zhuo, S.~Wen, C.-H. Lee, and S.-H.~G. Chan,
  ``Run, don't walk: chasing higher flops for faster neural networks,'' in
  \emph{Proceedings of the IEEE/CVF conference on computer vision and pattern
  recognition}, 2023, pp. 12\,021--12\,031.

\bibitem{Zhang2018}
X.~Zhang, X.~Zhou, M.~Lin, and J.~Sun, ``Shufflenet: An extremely efficient
  convolutional neural network for mobile devices,'' in \emph{Proceedings of
  the IEEE conference on computer vision and pattern recognition}, 2018, pp.
  6848--6856.

\bibitem{Hu2018}
J.~Hu, L.~Shen, and G.~Sun, ``Squeeze-and-excitation networks,'' in
  \emph{Proceedings of the IEEE conference on computer vision and pattern
  recognition}, 2018, pp. 7132--7141.

\bibitem{Lim2017}
B.~Lim, S.~Son, H.~Kim, S.~Nah, and K.~Mu~Lee, ``Enhanced deep residual
  networks for single image super-resolution,'' in \emph{Proceedings of the
  IEEE conference on computer vision and pattern recognition workshops}, 2017,
  pp. 136--144.

\bibitem{Wang2024}
Y.~Wang and T.~Zhang, ``Osffnet: Omni-stage feature fusion network for
  lightweight image super-resolution,'' in \emph{Proceedings of the AAAI
  Conference on Artificial Intelligence}, vol.~38, no.~6, 2024, pp. 5660--5668.

\bibitem{Zamfir2024}
E.~Zamfir, Z.~Wu, N.~Mehta, Y.~Zhang, and R.~Timofte, ``See more details:
  Efficient image super-resolution by experts mining,'' in \emph{Forty-first
  International Conference on Machine Learning}, 2024.

\bibitem{Agustsson2017}
E.~Agustsson and R.~Timofte, ``Ntire 2017 challenge on single image
  super-resolution: Dataset and study,'' in \emph{Proceedings of the IEEE
  conference on computer vision and pattern recognition workshops}, 2017, pp.
  126--135.

\bibitem{Bevilacqua2012}
M.~Bevilacqua, A.~Roumy, C.~Guillemot, and M.-L.~A. Morel, ``Low-complexity
  single-image super-resolution based on nonnegative neighbor embedding,'' in
  \emph{British Machine Vision Conference (BMVC)}, 2012.

\bibitem{Zeyde2012}
R.~Zeyde, M.~Elad, and M.~Protter, ``On single image scale-up using
  sparse-representations,'' in \emph{Curves and Surfaces: 7th International
  Conference, Avignon, France, June 24-30, 2010, Revised Selected Papers
  7}.\hskip 1em plus 0.5em minus 0.4em\relax Springer, 2012, pp. 711--730.

\bibitem{Martin2001}
D.~Martin, C.~Fowlkes, D.~Tal, and J.~Malik, ``A database of human segmented
  natural images and its application to evaluating segmentation algorithms and
  measuring ecological statistics,'' in \emph{Proceedings eighth IEEE
  international conference on computer vision. ICCV 2001}, vol.~2.\hskip 1em
  plus 0.5em minus 0.4em\relax IEEE, 2001, pp. 416--423.

\bibitem{Huang2015}
J.-B. Huang, A.~Singh, and N.~Ahuja, ``Single image super-resolution from
  transformed self-exemplars,'' in \emph{Proceedings of the IEEE conference on
  computer vision and pattern recognition}, 2015, pp. 5197--5206.

\bibitem{Matsui2017}
Y.~Matsui, K.~Ito, Y.~Aramaki, A.~Fujimoto, T.~Ogawa, T.~Yamasaki, and
  K.~Aizawa, ``Sketch-based manga retrieval using manga109 dataset,''
  \emph{Multimedia tools and applications}, vol.~76, pp. 21\,811--21\,838,
  2017.

\bibitem{Zhao2020}
H.~Zhao, X.~Kong, J.~He, Y.~Qiao, and C.~Dong, ``Efficient image
  super-resolution using pixel attention,'' in \emph{Computer Vision--ECCV 2020
  Workshops: Glasgow, UK, August 23--28, 2020, Proceedings, Part III 16}.\hskip
  1em plus 0.5em minus 0.4em\relax Springer, 2020, pp. 56--72.

\bibitem{Sun2023}
L.~Sun, J.~Dong, J.~Tang, and J.~Pan, ``Spatially-adaptive feature modulation
  for efficient image super-resolution,'' in \emph{Proceedings of the IEEE/CVF
  International Conference on Computer Vision}, 2023, pp. 13\,190--13\,199.

\bibitem{Wang2024b}
Y.~Wang, Y.~Li, G.~Wang, and X.~Liu, ``Multi-scale attention network for single
  image super-resolution,'' in \emph{Proceedings of the IEEE/CVF Conference on
  Computer Vision and Pattern Recognition}, 2024, pp. 5950--5960.

\bibitem{Wu2024}
G.~Wu, J.~Jiang, J.~Jiang, and X.~Liu, ``Transforming image super-resolution: A
  convformer-based efficient approach,'' \emph{IEEE Transactions on Image
  Processing}, 2024.

\bibitem{Ni2025}
N.~Ni and L.~Zhang, ``Slvr: Super-light visual reconstruction via blueprint
  controllable convolutions and exploring feature diversity representation,''
  in \emph{Proceedings of the Computer Vision and Pattern Recognition
  Conference}, 2025, pp. 400--410.

\bibitem{Guo2024}
H.~Guo, J.~Li, T.~Dai, Z.~Ouyang, X.~Ren, and S.-T. Xia, ``Mambair: A simple
  baseline for image restoration with state-space model,'' in \emph{European
  conference on computer vision}.\hskip 1em plus 0.5em minus 0.4em\relax
  Springer, 2024, pp. 222--241.

\bibitem{Li2025}
B.~Li, H.~Zhao, W.~Wang, P.~Hu, Y.~Gou, and X.~Peng, ``Mair: A locality-and
  continuity-preserving mamba for image restoration,'' in \emph{Proceedings of
  the Computer Vision and Pattern Recognition Conference}, 2025, pp.
  7491--7501.

\bibitem{Chen2023a}
Z.~Chen, Y.~Zhang, J.~Gu, L.~Kong, X.~Yang, and F.~Yu, ``Dual aggregation
  transformer for image super-resolution,'' in \emph{Proceedings of the
  IEEE/CVF international conference on computer vision}, 2023, pp.
  12\,312--12\,321.

\bibitem{Gu2021}
J.~Gu and C.~Dong, ``Interpreting super-resolution networks with local
  attribution maps,'' in \emph{Proceedings of the IEEE/CVF Conference on
  Computer Vision and Pattern Recognition}, 2021, pp. 9199--9208.

\end{thebibliography}

%\begin{IEEEbiography}[{\includegraphics[width=1in,height=1.25in,clip,keepaspectratio]{tang.pdf}}]{Yinggan Tang}
%received the B.Sc. in Instrument Science, the M.Sc. and Ph.D degrees in Control Science and Engineering  in, respectively, 1999, 2002 and 2006 from Yanshan University, China. Currently,  he is a Professor in the Department of Automation at Yanshan University, China. His main research interests include image processing, system modeling and computational intelligence.
%\end{IEEEbiography}
%
%\begin{IEEEbiography}[{\includegraphics[width=1in,height=1.25in,clip,keepaspectratio]{hu.pdf}}]{Quanwei Hu}
%received his bachelor's degree in Mechanical Design and Manufacturing and Automation from Zhengzhou University of Light Industry in 2022. He is currently pursuing his Master's degree in Control Science and Engineering at Yanshan University. His research interests include image super-resolution and deep learning.
%\end{IEEEbiography}
%
%\begin{IEEEbiography}[{\includegraphics[width=1in,height=1.25in,clip,keepaspectratio]{bu.pdf}}]{Chunning Bu}
%received the B.E degrees in Electrical Engineering and Automation in 2014 from Shijiazhuang University and the M.Sc.degrees in Control  Engineering in 2017 from Yanshan University, China. Currently, he is a lecturer in the Department of Automation at Cangzhou Jiaotong College, China. His main research interests include image processing, electromagnetic interference and Multi-objective optimization.
%\end{IEEEbiography}

\end{document}